\documentclass[sigconf, natbib=false]{acmart}
\RequirePackage{filecontents}
\PassOptionsToPackage{dvipsnames}{xcolor}
\RequirePackage{comgipp}

\RequirePackage{amsfonts}
\RequirePackage{amsmath}

\usepackage{stmaryrd}
\usepackage{trimclip}
\usepackage{stackengine}
\usepackage{scalerel}
\usepackage{pgfplots}
\pgfplotsset{compat=newest}%
\usepackage{booktabs}
\usepackage{array}
\usepackage{wasysym}
\usepackage{cleveref}
\usepackage{adjustbox}
\usepackage{url}
\usepackage{pifont}
\usepackage{appendix}

\usepackage{xcolor}
\usepackage{CJKutf8}
\usepackage{siunitx} 
\usepackage{amsmath}

\usepackage{makecell}

\newcolumntype{P}[1]{>{\raggedright\arraybackslash}p{#1}}
\usepackage[flushleft]{threeparttable}

\makeatletter
\DeclareRobustCommand{\shortto}{%
 {\mathrel{\mathpalette\short@to\relax}}%
}

\newcommand{\short@to}[2]{%
 \mkern2mu
 \clipbox{{.5\width} 0 0 0}{$\m@th#1\vphantom{+}{\rightarrow}$}%
}
\newcommand{\shorttorhd}{%
  {\hstretch{0.6}{\gg}}
}

\renewcommand{\ceur}{EEKE 2021 - Workshop on Extraction and Evaluation of Knowledge Entities from Scientific Documents}
\renewcommand{\ceurLicense}{Copyright 2021 for this paper by its authors. Use permitted under Creative Commons License Attribution 4.0 International (CC BY 4.0).}

\setcopyright{none}
\renewcommand\footnotetextcopyrightpermission[1]{}
\setcopyright{rightsretained}
\copyrightyear{2021}
\acmYear{2021}
\acmDOI{10.1145/nnnnnnn.nnnnnnn}

\acmConference[EEKE '21]{2nd Workshop on Extraction and Evaluation of Knowledge Entities from Scientific Documents}{September 27--30, 2021}{Online}
\acmBooktitle{2nd Workshop on Extraction and Evaluation of Knowledge Entities from Scientific Documents, September 27--30, 2021, Online}
\acmPrice{15.00}
\acmISBN{978-1-4503-XXXX-X/18/06}

\usepackage[
	backend=biber,
	style=numeric,
	firstinits=true,
	url=false,
	isbn=false,
	]{biblatex}
\newcommand{\theReferenceText}{\fullcite{Stegmueller2021}}
\begin{document}

\definecolor{darkgreen}{RGB}{0,127,127}
\definecolor{closa}{RGB}{228,26,28}
\definecolor{conceptnet}{RGB}{55,126,184}
\definecolor{useml}{RGB}{152,78,163}
\definecolor{clesa}{RGB}{77,175,74}
\definecolor{clasa}{RGB}{102, 102, 102}

\title[Detecting Cross-Language Plagiarism]{Detecting Cross-Language Plagiarism using~Open~Knowledge~Graphs}

\author{Johannes Stegmüller}
\authornotemark[1]
\email{stegmueller@gipplab.org}
\orcid{0000-0001-5080-1808}
\affiliation{%
  \institution{University of Wuppertal}
  \city{Wuppertal}
  \country{Germany}
}

\author{Fabian Bauer-Marquart}
\authornotemark[1]
\email{fabian.marquart@uni-konstanz.de}
\affiliation{%
  \institution{University of Konstanz}
  \city{Konstanz}
  \country{Germany}
}

\author{Norman Meuschke}

\email{meuschke@uni-wuppertal.de}
\orcid{0000-0003-4648-8198}
\affiliation{%
  \institution{University of Wuppertal}
  \city{Wuppertal}
  \country{Germany}
}

\author{Terry Ruas}
\email{ruas@uni-wuppertal.de}
\orcid{0000-0002-9440-780X}
\affiliation{%
  \institution{University of Wuppertal}
  \city{Wuppertal}
  \country{Germany}
}

\author{Moritz Schubotz}
\email{moritz.schubotz@fiz-karlsruhe.de}
\orcid{0000-0001-7141-4997}
\affiliation{%
  \institution{FIZ Karlsruhe}
  \city{Berlin}
  \country{Germany}
}

\author{Bela Gipp}
\email{gipp@uni-wuppertal.de}
\orcid{0000-0001-6522-3019}
\affiliation{%
  \institution{University of Wuppertal}
  \city{Wuppertal}
  \country{Germany}
}

\renewcommand{\shortauthors}{Stegmüller, et al.}
\fancyhead{}
\begin{abstract}

Identifying cross-language plagiarism is challenging, especially for distant language pairs and sense-for-sense translations. We introduce the new multilingual retrieval model Cross-Language Ontology-Based Similarity Analysis (CL\nobreakdash-OSA) for this task. CL-OSA represents documents as entity vectors obtained from the open knowledge graph Wikidata. Opposed to other methods, CL\nobreakdash-OSA does not require computationally expensive machine translation, nor pre-training using comparable or parallel corpora. It reliably disambiguates homonyms and scales to allow its application to Web-scale document collections. We show that CL-OSA outperforms state-of-the-art methods for retrieving candidate documents from five large, topically diverse test corpora that include distant language pairs like Japanese-English. For identifying cross-language plagiarism at the character level, CL-OSA primarily improves the detection of sense-for-sense translations. For these challenging cases, CL-OSA's performance in terms of the well-established PlagDet score exceeds that of the best competitor by more than factor two. The code and data of our study are openly available.
\end{abstract}




\settopmatter{printfolios=true}
\sloppy
\maketitle

\thispagestyle{firststyle}

\section{Introduction} \label{sec.introduction}
\let\thefootnote\relax\footnotetext{*Both lead authors contributed equally to this research and writing the paper.}
Plagiarism is ``the use of ideas, concepts, words, or structures without appropriately acknowledging the source to benefit in a setting where originality is expected'' \cite{Fishman09}. Plagiarism harms scientific discourse, wastes resources, and can unjustifiably benefit the plagiarist if it remains undiscovered \cite[p.~22ff.]{WeberWulff14}. If researchers revise earlier results in later publications, papers that plagiarized the original findings remain unchanged. Others may spend time and resources trying to replicate such wrong results, or worse, consider them correct and compromise later research or practical applications. Reviewing and sanctioning plagiarized research papers or grant applications often require hundreds of working hours from the reviewers, affected academic institutions, and funding agencies. 

The rapid advancement of the Web and information technology have enabled convenient access to vast amounts of information, making plagiarism easier than ever. This development has spurred extensive research on automated methods to identify plagiarized content. Most state-of-the-art plagiarism detection methods analyze lexical, syntactic, and semantic text similarity to identify copied or moderately obfuscated monolingual plagiarism \cite{FoltynekMG19}. 

Detecting cross-language plagiarism remains a significant challenge, despite advances in cross-language information retrieval (CLIR) \cite{WeberWulff14, FoltynekMG19}. Most current cross-language plagiarism detection (CLPD) methods (cf. \Cref{sec.related_work}) rely on computationally expensive machine translation or learning approaches based on parallel or comparable corpora that are not easily available for many languages. Thus far, few detection methods leverage multilingual knowledge graphs to analyze the deep semantic similarity of documents. This is one of the reasons why current methods can only identify mildly obfuscated cross-language plagiarism reliably \cite{FoltynekMG19}.

To fill this gap, we propose a new multilingual retrieval model and apply it to CLPD. The main contributions of our work are:
 \begin{enumerate}
     \item We introduce Cross-Language Ontology-Based Similarity Analysis as a novel CLPD method. CL\nobreakdash-OSA identifies the semantic similarity of documents by leveraging multilingual knowledge graphs like Wikidata\footnote{\url{https://wikidata.org}} to extract and compare entities contained in the documents. It models texts as entity vectors and leverages relations between entities for entity disambiguation. CL-OSA is suitable for all topical domains, robust against paraphrasing, and applicable to many close and distant language pairs. 
     
     \item Using documents in Chinese, French, English, Japanese, and Spanish, we show that CL\nobreakdash-OSA outperforms state-of-art methods for the two standard sub-tasks in CLPD\textemdash candidate retrieval and detailed analysis.
     
     \item We make our source code and data publicly available.
     
 \end{enumerate}


\section{Related Work} \label{sec.related_work}

Cross-language plagiarism detection is an information retrieval task that methods typically address in two steps \cite{PotthastBSR11}. In the \textit{candidate retrieval} step, the methods use efficient algorithms to retrieve from a large document collection in another language (\textit{reference collection}) all documents that contain a certain amount of similar content as the input document. In the \textit{detailed analysis} step, the methods perform pairwise comparisons of the input document to each candidate to identify similar segments within the documents at the character level. Hereafter, we summarize CLPD and general cross-language information retrieval approaches relevant to our work.
\subsection{Machine Translation}
Many CLIR and CLPD methods combine language normalization via machine translation with monolingual similarity analysis \cite{FoltynekMG19, MeuschkeG13}. Cross-language Character $n$-grams (CL\nobreakdash-CNG) proposed by McNamee and Mayfield \cite{McNameeM04} is a vector space retrieval model that uses machine translation to map two documents into a common language, typically English. The method then partitions both documents into character $n$\nobreakdash-grams exclusively consisting of lowercase letters and numbers. CL\nobreakdash-CNG computes the cosine measure for the $n$\nobreakdash-gram vectors to determine their similarity. Several studies on CLPD use CL\nobreakdash-CNG as a baseline approach, e.g., \cite{PotthastBSR11,BarronCedenoGR13,FrancoSalvadorGR13,FrancoSalvadorGRB16}.

Chen et al. \cite{ChenZZ05} combined machine-translation with a vector space model (VSM) for ranked cross-language retrieval of documents in English and Chinese. Their method translates the query using a bilingual dictionary before performing ranked retrieval using the VSM. Their study showed that segmenting Chinese texts is challenging for achieving high retrieval quality when the query is in another language. Franco-Salvador et al. used a similar approach as a baseline in their evaluation \cite{FrancoSalvadorGRB16}. Their Cross-language Vector Space Model (CL\nobreakdash-VSM) represents documents in a bilingual form by concatenating \textit{tf\nobreakdash-idf}\nobreakdash-weighted vector representations of the original document and its translation obtained using a statistical dictionary. The authors re-weighted the vector representing the translated document using the translation probabilities of words.

Barrón-Cedeño et al. \cite{BarronCedenoRPJ08} proposed Cross-Language Alignment-based Similarity Analysis (CL\nobreakdash-ASA) for the CLPD task. The method uses statistical machine translation based on the \textit{IBM alignment model 1}~\cite{BrownCPP90}. In a later study performed by the same research group, CL\nobreakdash-ASA achieved superior precision over CL\nobreakdash-CNG, which achieved the highest recall \cite{BarronCedenoGR13}. CL\nobreakdash-ASA is more robust against synonym replacements than CL\nobreakdash-CNG because it considers multiple translation candidates and their translation probabilities. However, this approach also causes CL\nobreakdash-ASA to be computationally more expensive than CL\nobreakdash-CNG. CL\nobreakdash-ASA requires computing the similarities between all documents, while CL\nobreakdash-CNG is typically implemented using an index, thereby achieving faster query execution.

\subsection{Corpus-based Semantics}
Corpus-based semantic analysis follows the idea of distributional semantics, i.e., words co-occurring in similar contexts tend to convey similar meaning. Consequently, one assumes that texts with similar word distributions are semantically similar \cite{GomaaF13}. Word embeddings and Semantic Concept Analysis (SCA) are established corpus-based semantic analysis approaches that researchers applied to CLIR and CLPD, besides many other tasks. The approaches differ in the scope within which they consider co-occurring words. 

\subsubsection{Word Embeddings}\label{sec.rw.word_embeddings}
Word embeddings consider the surrounding words to represent a word in a dense, low-dimensional, fixed-size vector space. Words with similar neighboring words should be close to each other in the vector space \cite{BengioDVJ03}. 

Ferrero et al. proposed two CLPD methods based on word embeddings \cite{FerreroABS17}. The first, Cross-Language Conceptual Thesaurus-based
Similarity Word Embedding (CL-CTS-WE), represents a word as a bag of words (BOW) consisting of the 10 most similar words according to the embeddings model. The second, Cross-Language Word Embedding Sentence Vector (CL-WES), represents sentences as the sum of the embedding vectors of their constituent words and compares the resulting sentence vectors using the cosine measure. Both methods use Multivec \cite{BerardSPB16} as their pre-trained word embeddings model. Multivec combines word2vec \cite{MikolovSCC13}, paragraph vectors \cite{LeM14}, and bilingual distributed representations \cite{LuongPM15}.

Glava{\v s} et al. presented a computationally lightweight method to analyze cross-language similarity for language pairs that lack parallel corpora or named entity recognition \cite{GlavasFPR18}. The authors mapped words into a bilingual embedding space by initially creating a monolingual word embedding and then applying a linear function learned from a training corpus.

In our evaluation (cf. \Cref{sec.evaluation}), we use ConceptNet and USE-ML, which are comparable to the methods Ferrero et al. \cite{FerreroABS17} and Glava{\v s} et al. \cite{GlavasFPR18} proposed, but rely on more recent pre-trained word embedding models. Different from Ferrero et al. \cite{FerreroABS17}, we represent the documents in our datasets as the average of their constituent word embeddings from the pre-trained models.

ConceptNet-Numberbatch (ConceptNet) \cite{SpeerCH17} uses traditional word embeddings, such as word2vec \cite{MikolovSCC13} and GloVe \cite{PenningtonSM14}, and the lexical information in ConceptNet\footnote{\url{http://conceptnet.io/}} to derive its semantic vectors.

The Universal Sentence Encoder-Multilingual (USE-ML) \cite{YangCAG20} offers two architectures to derive its vectors. One is inspired by the Transformer architecture \cite{VaswaniSPU17} and the other uses Deep Average Networks (DAN) \cite{IyyerMBD15}.

\subsubsection{Semantic Concept Analysis}
Semantic concept analysis extends the distributional semantics idea to an external corpus.

Potthast et al. \cite{PotthastSA08} introduced Cross-Language Explicit Semantic Analysis (CL\nobreakdash-ESA) as a multilingual generalization of the semantic retrieval model Explicit Semantic Analysis (ESA) proposed by Gabrilovich and Markovitch \cite{GabrilovichM07}. ESA and CL\nobreakdash-ESA represent documents as vectors in a high-dimensional vector space of semantic concepts, which are explicitly encoded topics in a knowledge base corpus. CL\nobreakdash-ESA uses a concept-aligned comparable corpus available in multiple languages. Specifically, Potthast et al. used Wikipedia articles and considered each article available in multiple languages to represent one concept. Each dimension of a document vector represents the \textit{tf\nobreakdash-idf} similarity of the document to one of the concepts. The similarity of document vectors is typically quantified using the cosine measure \cite{GabrilovichM07,PotthastSA08}. Meuschke et al. extended CL-ESA by also considering the order in which concepts occur in the text to identify potentially suspicious patterns \cite{MeuschkeSSG17}.

The evaluations of Potthast et al. \cite{PotthastSA08} showed that CL\nobreakdash-ESA performs best if the concept space has 100,000 or more dimensions, i.e., if at least 100,000 Wikipedia articles are considered. In this case, CL\nobreakdash-ESA achieved a recall above 0.90 for the JRC\nobreakdash-Acquis corpus \cite{SteinbergerPWI06}. However, such high dimensionality is computationally expensive. Therefore, Potthast et al. advised that: ``If high retrieval speed or a high multilinguality is desired, documents should be represented as 1~000\nobreakdash-dimensional concept vectors. At a lower dimension the retrieval quality deteriorates significantly. A reasonable trade-off between retrieval quality and runtime is achieved for a concept space dimensionality between 1~000 and 10~000.'' \cite[p.~526f.]{PotthastSA08}. 

Despite limitations in dimensionality as proposed by Potthast et al., CL\nobreakdash-ESA is computationally more expensive than CL\nobreakdash-ASA because it requires computing the similarity of the input document to all concepts followed by calculating the similarity of all document vector pairs as is the case for CL\nobreakdash-ASA.

\subsection{Non-textual Content Analysis}
Researchers proposed analyzing non-textual content features to overcome the ambiguities of natural language and complement text analysis approaches to improve the detection of concealed plagiarism forms, such as translations. The investigated content elements include academic citations \cite{GippMB14,MeuschkeSSG18,MazovGK16}, images \cite{EisaSA17ICCD,EisaSA20,MeuschkeGSB18}, and mathematical content \cite{SchubotzTSM19,MeuschkeSSK19,MeuschkeSHS17}. 

\subsection{Knowledge-based Semantics}

Knowledge-based semantic analysis approaches, such as the one we propose, use entities encoded in semantic networks, such as thesauri, ontologies, and knowledge graphs.

Cross-Language Knowledge Graph Analysis (CL-KGA) is a CLPD method proposed by Franco-Salvador et al. and most related to our work \cite{FrancoSalvadorGR13}. CL\nobreakdash-KGA uses sub-graphs of the multilingual semantic network \textit{BabelNet}\footnote{\url{https:/babelnet.org}} to represent text segments. Specifically, CL\nobreakdash-KGA splits documents into segments using a five-sentences-long sliding window with a two-sentences step width, lemmatizes the segments, and performs part-of-speech tagging. By mapping terms in the preprocessed text segments to BabelNet, CL-KGA obtains the sub-graph of BabelNet used to represent the segments. Franco-Salvador et al. proposed a graph-based similarity measure that considers the similarity of entities and their relations to compare the entity representations. The authors improved the weighting function in subsequent publications and combined the graph-based representation with neural text representations \cite{FrancoSalvadorGR14, FrancoSalvadorGRB16, FrancoSalvadorRM16}.

\subsection{Neural Networks}
Neural text representations and language models have significantly advanced the state of the art for many NLP and CLIR tasks. A comprehensive review would exceed the scope of this paper; therefore, we restrict our description to successful neural CLIR methods, which we use as baselines for our evaluation in \Cref{sec.evaluation}.

The External-data Composition Neural Network (XCNN) is a cross-language continuous space model created by a composition function on top of a deep neural network. In difference to similar approaches, XCNN can be initialized with monolingual data and extended with at least a small set of parallel data. This feature of the network is especially useful for low-resource languages \cite{GuptaBR17}.

The Siamese Neural Network (S2Net) trains two identical networks concurrently with parallel data that has to be annotated with a similarity score \cite{FrancoSalvadorGRB16}. The network lends itself for similarity learning in a bilingual use case, where each network reflects data in one of the two languages. For each text input, the networks emit feature vectors representing the input in the respective languages, which can then be compared using the cosine similarity. 

Bilingual Autoencoders (BAE) are trained using bag-of-words representations of multiple sentences from parallel corpora as input. From the BOW representation in a source language, the encoder creates a BOW representation in the target language. During training, the encoder is optimized by minimizing the reconstruction error between the created representation from the source language and the original target representation \cite{GuptaBBC14}.

\subsection{Research Gap} 
Most CLPD methods rely on machine translation \cite{McNameeM04,ChenZZ05,BarronCedenoRPJ08,BarronCedenoGR13} or representations trained using parallel \cite{GuptaBR17,FrancoSalvadorGRB16} or comparable corpora \cite{GuptaBR17,PotthastSA08}. These approaches depend on lexical and syntactical similarity and topical homogeneity of the documents in different languages. There is a need for CLPD methods that can analyze a wide variety of topics across academic disciplines. The use of knowledge graphs has been shown to benefit the analysis of semantic document similarity in the monolingual and cross-language setting. However, few studies have investigated the use of knowledge graphs for cross-language plagiarism detection \cite{FrancoSalvadorGR13,FrancoSalvadorGR14, FrancoSalvadorGRB16, FrancoSalvadorRM16}. We extend and improve upon this prior work, as we explain hereafter.

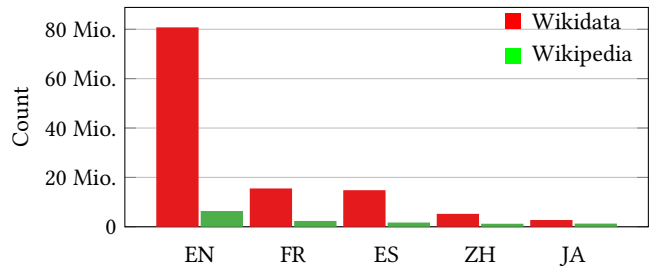
\begin{figure}[t!]
\begin{tikzpicture}[
          blacknode/.style={shape=circle, draw=black, line width=2},
          bluenode/.style={shape=circle, draw=blue, line width=2},
          greennode/.style={shape=rectangle, draw=green, fill=green, line width=2},
          rednode/.style={shape=rectangle, draw=red, fill=red, line width=2}
        ]
        \begin{axis}[
            symbolic x coords={EN, FR, ES, ZH, JA},
            width  = 0.48*\textwidth,
            height = 4.5cm,
            major x tick style = transparent,
            ybar=2*\pgflinewidth,
            bar width=16pt,
            ymajorgrids = true,
            ylabel = {Count},
            ymajorgrids = true,
            xtick = data,
            ytick = {0, 20000000, 40000000, 60000000, 80000000},
            yticklabels = {0, 20 Mio., 40 Mio., 60 Mio., 80 Mio.},
            scaled y ticks = false,
            enlarge x limits=0.2,
            ymin=0,
            legend cell align=left,
            legend style={
                    at={(1,1.05)},
                    anchor=south east,
                    column sep=1ex
            }
        ]
        \addplot[ybar, mark=no, fill=closa, draw=none] plot coordinates {
            (EN, 80 767 852)
            (FR, 15 520 782)
            (ES, 14 813 859)
            (ZH, 5 232 869)
            (JA, 2 735 750)
         };
        \addplot[ybar, mark=no, fill=clesa, draw=none] plot coordinates {
            (EN, 6 382 074)
            (FR, 2 362 216)
            (ES, 1 716 529)
            (ZH, 1 231 077)
            (JA, 1 291 697)
         };
        \end{axis}
        \matrix [draw,below left, draw opacity=0.0] at (current bounding box.north east) {
              \node [rednode,label=right:Wikidata] {}; \\
              \node [greennode,label=right:Wikipedia] {}; \\
        };

    \end{tikzpicture}
    \caption{Number of Wikidata entities (red) and Wikipedia articles (green) per language as of September 2021.}
\label{tab.relwork.wikidata}
\end{figure}

\section{Proposed Method}
\label{sec.closa}

\begin{figure*}[t!]
\centering
\includegraphics[width=0.9\textwidth]{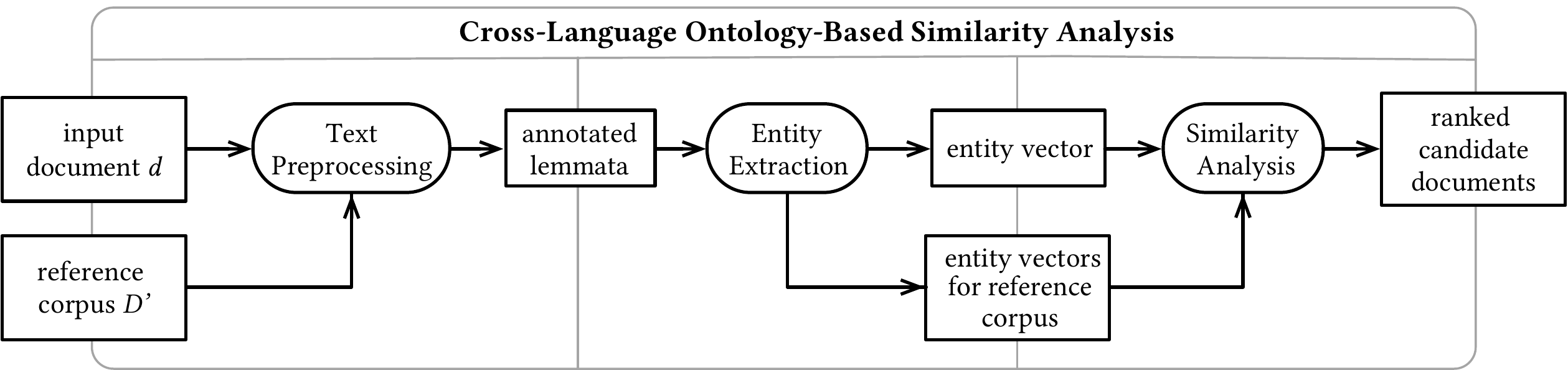}
\caption{Overview of Cross-Language Ontology-based Similarity Analysis (CL-OSA).}
\label{fig.closa_overview}
\end{figure*}

Cross-Language Ontology-based Similarity Analysis is a multilingual retrieval model derived from a knowledge graph that includes ontological relations. The method constructs language-independent, semantically-enhanced entity vectors that not only include entities present in the modeled documents but also entities that are hierarchically linked by \textsl{subclass of} and \textsl{instance of} relations. 

Three reasons governed our decision to use the open knowledge graph Wikidata to realize CL-OSA, instead of an open encyclopedia like Wikipedia used, e.g., by ESA and CL-ESA, or BabelNet used by CL-KGA. First, the number of entities in Wikidata greatly exceeds the number of Wikipedia articles. ESA and CL-ESA use Wikipedia articles as concepts, which limits their representation. \Cref{tab.relwork.wikidata} shows the number of Wikidata entities per language (queried from an official JSON dump dated September 2021). There are more than twelve times as many Wikidata entities available for English as there are Wikipedia articles. For Spanish and French, the number of Wikidata entities exceeds the number of Wikipedia articles by factors between six and eighth. For Chinese the factor is four and for Japanese the factor is two.

Second, while Wikipedia exclusively contains cross-references between articles, Wikidata includes property links that express relationships, such as \textsl{instance of}, \textsl{subclass of}, \textsl{color}, or \textsl{part of}. Therefore, Wikidata offers a wider range of typed relationships that are readily accessible for automated processing. 

Third, Wikidata offers public domain data with no restrictions on its use. BabelNet, for example, imposes fees for commercial use\footnote{\url{https://babelnet.org/full-license} | \url{https://babelscape.com/wordatlas}}.

As \Cref{tab.relwork.wikidata} shows, both the number of Wikipedia articles and Wikidata entities differs greatly between languages. Fewer entities can reduce the detection effectiveness of knowledge-graph-based detection methods like CL-OSA for the respective language. However, as we show in our evaluation (cf. \Cref{sec.evaluation}) CL-OSA and comparable methods already achieve good results for languages with fewer entities, like Japanese and Chinese. Moreover, knowledge bases like Wikidata grow continuously, especially due to significant advances in automated entity extraction and linking \cite{AlMoslmiGLV20}.

\subsection{Retrieval Model} \label{sec.model}
\Cref{fig.closa_overview} illustrates the three-step process consisting of \textit{Text Preprocessing}, \textit{Entity Extraction}, and \textit{Similarity Analysis} CL-OSA follows for representing documents as language-independent entity vectors and using them for ranked cross-language retrieval. Hereafter, we formalize the process and present each of its three steps in detail. 

Let $D$ be the set of (suspicious) input documents and $D'$ be the set of documents in the reference collection, i.e., potential sources for content. The goal is to determine the similarity between an input document $d \in D$ and a candidate document $d' \in D'$ denoted as $\varphi(d, d')$. CL-OSA represents a document $d = w_1 w_2 \dots w_k$ written in language $L_1 \in \mathcal{L}$ as an hierarchy-enhanced entity vector $\mathbf{d}_\shorttorhd$.

\[ \mathbf{d}_\shorttorhd = ( q_1, \dots, q_n ).\] 

The elements of $\mathbf{d}_\shorttorhd$ are entities of the knowledge graph occurring in the document and the ontological ancestors of these entities.

A multilingual knowledge graph $Q$ such as Wikidata is a set of entities $q$, defined as a tuple $q = (\Sigma, \mathcal{A}, \Delta, \to)$ where
\begin{itemize}
	\item $\Sigma$ is a set of labels $\{l_1, \dots, l_m\}$,
	\item $\mathcal{A}$ is a set of alias sets $\{A_1, \dots, A_m\}$,
	\item $\Delta$ is a set of descriptions $\{ \delta_1, \dots, \delta_m \}$, and
	\item $\to \subseteq Q \times P \times 2^Q$ is a property relation.
\end{itemize}

$P \subsetneq Q$, $\mathcal{L} = \{ L_1, \dots, L_m \}$, and $\mathcal{T}$ denote sets of properties, languages, and topics respectively. 

For example, the entity \textit{query} has the English label \textit{query}, the English alias \textit{database query}, the description \textit{precise request for information retrieval}, and the \textit{subclass of} property that maps \textit{query} to the entity for \textit{information request}, all of which is valuable information when relating entities to each other.

For convenience, we denote property relations as $q \xrightarrow{p} q'$ instead of $(q, p, q') \in \to$ using the following notations:
\begin{itemize}
    \item $p_\shortto$ denotes the \textsl{instance of} relation
    \item $p_\rhd$ denotes the \textsl{subclass of} relation
    \item $p_\shorttorhd$ the \textit{combined} property relation such that\\$q \xrightarrow{p_\shorttorhd} q' = q \xrightarrow{p_\shortto} q' \lor  q \xrightarrow{p_\rhd} q'$
    \item $\to^*$ denotes the \textit{transitive} property relation, which we define as $q \xrightarrow{\smash{p}}^* q' = q \xrightarrow{p_\shorttorhd} q_1 \xrightarrow{p_\shorttorhd} \dots \xrightarrow{p} q'$
\end{itemize}

The \textit{transitive} property relation lets child entities inherit their parents' properties. For example, the entity \textit{pi} is an \textit{instance of} \textit{mathematical constant}, which itself is a \textit{subclass of} \textit{number}. Therefore, \textit{pi} is transitively also an \textit{instance of} \textit{number}.

\subsection{Text Preprocessing}
\label{textpreprocessing}

The goal of the preprocessing is to avoid typical issues that arise if statistical machine translation or alignment-based retrieval models are employed for the language normalization step of the CLPD process. Translations produced using these state-of-the-art methods often exhibit grammatical errors, syntactical differences, and sub-optimal wording if sufficient domain-specific training data is missing. The availability of training data is often problematic for highly domain-specific texts, such as scientific, technical, and professional documents in languages other than English.

As an example to illustrate these challenges, we use the introductory sentence of the article \textit{Volkswirtschaftslehre} (macroeconomics) in the German Wikipedia\footnote{\url{https://de.wikipedia.org/wiki/Volkswirtschaftslehre}}: 

\begin{quote}\label{quote.VWL}
\textit{Die Volkswirtschaftslehre (auch Nationalökonomie oder wirtschaftliche Staatswissenschaften kurz VWL) ist ein Teilgebiet der Wirtschaftswissenschaft.}
\end{quote}

Translating this sentence to English using Google Translate introduces ambiguity by mapping several words that have a clear distinction in German to ``economics``: 
\begin{quote}\textit{Economics (also economics or economic political science short VWL) is a branch of economics.}
\end{quote}

To avoid these issues, CL-OSA first categorizes the document by topic and then extracts terms from the document corresponding to Wikidata entities having the most ancestors. The idea is that in this way, CL-OSA selects entities that are most specific because they have many superordinate entities and are thus representative of the document's topic. Including superordinate concepts also helps to disambiguate entities in the text, since the included superordinate (potentially ambiguous) entities likely also occur in the document. 

Specifically, CL-OSA performs the following steps to prepare documents for being represented as entity vectors. 

\textbf{Language Detection.} In case the language of a document is unknown, CL-OSA infers it using a language detector\footnote{\url{https://github.com/optimaize/language-detector}}. The detector uses the vector space model to compare a document's $n$-grams to a pre-trained set of $n$-grams from various multilingual comparable corpora, and selects the most probable language.

\textbf{Topic Detection.} To roughly identify the topical domain of a document, we trained a Bayesian classifier using approximately 100 bag-of-words 
representations of Wikipedia articles that we hand-assigned to topics relevant for our entity extraction and disambiguation approaches, i.e., \textit{biology}, \textit{fiction}, and \textit{neutral}. Classifying a document yields the topic that has the highest word overlap with the topic-specific articles on which the classifier has been trained. Using only these three categories was sufficient because they help to disregard the largest amount of Wikidata entities that are irrelevant for the extraction task, e.g., movie titles for works not related to fiction, or genes and proteins for works not related to biology.

\textbf{Tokenization and Word Segmentation.} Depending on the document's language, CL-OSA performs tokenization or word segmentation to split the text into a token sequence. Our method uses simple tokenization for white-space separated languages, such as English, Korean, or French, and employs more sophisticated methods, such as a dictionary lookup, for languages lacking a word delimiter, e.g., Chinese or Japanese. The tokenization step keeps stop-words, but strips punctuation.

\textbf{Lemmatization.} To exploit entity labels and aliases, which Wikidata contains in their base forms, our method lemmatizes derived and inflected tokens. Additionally, it uses WordNet \cite{Fellbaum98} for mapping verbs to nouns. A fallback procedure if no lemmatizer is available for a language would be using a rule-based stemmer, although this could introduce ambiguity. Therefore, we did not employ stemming. Skipping lemmatization is safe for languages that lack inflection or for which the tokenization step performs inflection removal, e.g., Chinese and Japanese.

\textbf{Named Entity Recognition.} To reduce the ambiguity of tokens, CL-OSA performs part-of-speech (POS) tagging and named entity (NE) recognition to annotate the lemmata with POS and proper noun information, such as \textit{location}, \textit{human}, or \textit{organization}.

For tokenization, lemmatization, and named entity extraction, we use Stanford CoreNLP \cite{ManningSBF14} for European languages and Kuromoji\footnote{\url{https://www.atilika.org}} for Chinese and Japanese. 

\subsection{Entity Extraction} \label{sec.entity_extraction}
\begin{table}
\caption{Entity extraction and disambiguation for a text fragment from an editorial in the English Financial Times.}
\small
\begin{tabular*}{0.48\textwidth}{p{1.9cm} p{0.5cm} p{0.5cm} p{0.8cm} p{3.2cm}}
\toprule
\textbf{Lemma} & \textbf{POS} & \textbf{NE} & \textbf{Entity} & \textbf{English label $l_{\text{en}}$}
\\
\midrule
US							& NNP	& LOC 		& $\underline{q_{30}}$ 			& United States of America \\
							&		&				& $q_{64142888}$				& User Systems (United States)\\
tax 						& NN 	& O 			& $q_{8161}$ 					& tax\\
authorities 				& NNS 	& O				& $\underline{q_{174834}}$ 		& authority\\ 
							&		&				& $q_{13424378}$ 				& authority [rulership]\\
							&		&				& $q_{59646503}$ 				& authority record\\
							&		&				& $q_{15708736}$ 				& public authority\\
teeth 						& NNS 	& O				& $q_{55347892}$ 				& tooth [heraldic]\\
	 						&		&				& $q_{47450777}$ 				& tooth [gear]\\
							&		&				& $\underline{q_{553}}$			& tooth [jaw]\\
							&		&				& $q_{15043709}$				& tine\\
battle 						& NN 	& O 			& $q_{4869972}$ 				& battle [medieval]\\
	 						&		&				& $\underline{q_{178561}}$ 		& battle\\
politicians 				& NNS 	& O 			& $q_{82955}$ 					& politician\\
Internal Revenue Service 	& NNP 	& ORG	& $q_{973587}$ 					& Internal Revenue Service\\
appears be 					& VBZ 	& O				& $\underline{q_{3620816}}$ 	& appearance\\
							&		&				& $q_{4207474}$ 				& semblance\\
stance 						& NN	& O				& $\underline{q_{172378}}$ 		& stance [martial arts]\\
							&		&				& $q_{7598021}$					& stance [football]\\
							&		&				& $q_{48302332}$ 				& stance [tennis]\\
							&		&				& $q_{17052364}$ 				& stance [linguistics]\\
international 				& JJ 	& O				& $\underline{q_{1072012}}$ 	& international\\
							&		&				& $q_{61029267}$			 	& rest of the world\\
leaves 						& VBZ 	& O				& $q_{24759450}$ 				& leave\\
							&		&				& $q_{19279529}$ 				& go \\
							&		&				& $\underline{q_{5338673}}$  	& annual leave\\
							&		&				& $q_{13561011}$ 				& leave of absence\\
taxpayers 					& NNS 	& O				& $q_{25211970}$ 				& taxpayer [building]\\
	 						&		&				& $\underline{q_{1938414}}$ 	& taxpayer\\
\bottomrule
\end{tabular*}
\label{tab.closa.example_extraction}
\end{table}
CL-OSA maps every lemma $n$-gram with $n \in \{1,2,3\}$ to entity candidates, which it obtains by querying the $n$-grams to the labels and aliases of the knowledge graph. We apply a coarse filter to disregard entities that are likely irrelevant depending on the document's topic. Specifically, we require that entity candidates for documents in the category \textit{fiction} are a subclass of or an instance of \textit{creative work}. For documents in the category \textit{biology}, entity candidates must be either a subclass of or an instance of \textit{gene}. Entity candidates originating from longer lemma $n$-grams take precedence over shorter sequences. Typically, CL-OSA retrieves multiple entity candidates, which it disambiguates as described in the following section.

\subsection{Entity Disambiguation}\label{sec.entity_disamb}
CL-OSA disambiguates entity candidates to a single entity using a combination of manually devised filters and mappings of topics to named entities. This procedure removes entities: 
\begin{itemize}
	\item if their original token has a POS tag related to punctuation, prepositions, symbols, markers, or personal pronouns
	\item that represent a stop-word such as ``and``
	\item that represent a Han character in case of Chinese or Japanese
	\item that represent a Wikimedia disambiguation page
	\item that are a subclass of \textit{natural number} and have numeric labels
	\item that are not an instance of their named entity types, e.g., \textit{human}, \textit{location}, and \textit{organization}
\end{itemize}
Additionally, we exploit the entity hierarchy by disambiguating to the entity candidate that has the most ontological ancestors contained in the text surrounding the entity candidate.

\Cref{tab.closa.example_extraction} illustrates the extraction and disambiguation of entities for the following text fragment taken from the ECCE corpus, which contains Financial Times editorials in English and Chinese \cite{Yang17}:

\begin{quote}
\textit{US tax authorities are finally finding their teeth. After a long battle with politicians, the Internal Revenue Service appears to be toughening its stance on international tax arbitrage that leaves taxpayers short-changed.}
\end{quote}

\Cref{tab.closa.example_extraction} shows the Wikidata entity candidates for every lemma-POS-NE triple extracted from the text. The table omits triples without entity candidates for brevity. The final entities after the disambiguation step are underlined.
\subsection{Similarity Analysis}
\label{sec.similarity_analysis}

In the final step, CL-OSA constructs the hierarchy-enhanced entity vector $\mathbf{d}_\shorttorhd$ by taking all entities $q$ from the bag-of-entities and applying boolean weighting, i.e., assigning a $1$ for entities that occur in the document, and a $0$ otherwise. Weighting using the raw term frequency yielded worse results in our experiments.

For example, when analyzing our example sentence from the German Wikipedia article \textit{Volkswirtschaftslehre} (cf. page \pageref{quote.VWL}), we obtain the entities \textit{general economics} and \textit{economics} for the German text. For the English text, we only obtain \textit{economics}. However, \textit{general economics} is part of \textit{economics}, and both are transitively instances of \textit{branch of science}. Therefore, as these entities are included in the vector, yet are weighted inversely proportional to their graph-based distance, the similarity score of both sentences increases without introducing too many commonalities.

Furthermore, CL-OSA leverages the relation $\smash{p}$ by adding all entities $q_a$ to $\mathbf{d}_\shorttorhd$ if they satisfy $\smash{p}^m(q) = q_a$ for any $m \in \{1,2,3\}$ and assigns the weight $(m+1)^{-2}$. That is, CL-OSA adds the ancestors of an entity to the vector and assigns an exponentially decreasing weight (inverse quadratic growth) the more distant the ancestors are. Thus, first-level ancestors get a weight of $1/2^2$, second-level ancestors $1/3^2$, and so forth. We derived this weighting from the similarity measure by Li et al. \cite{LiWZW13}, which has been shown to reflect semantic similarity in graphs. 

CL-OSA compares the resulting vector $\mathbf{d_\shorttorhd} = ( q_1, \dots, q_n )$ to all vectors $\mathbf{d'_\shorttorhd}$ in the reference collection $D'$ by computing the cosine similarity
\[\varphi(\mathbf{d_\shorttorhd}, \mathbf{d'_\shorttorhd}) = \frac{\mathbf{d_\shorttorhd} \cdot \mathbf{d'_\shorttorhd}}{||\mathbf{d_\shorttorhd}|| \, ||\mathbf{d'_\shorttorhd}||}\] and uses the scores to rank all reference documents $d' \in D'$ in decreasing order of their similarity to document $d$. 

\input{tables/tab.eval_cr.results}

\section{Evaluation} \label{sec.evaluation}
We evaluate CL-OSA's performance for the \textit{candidate retrieval} and \textit{detailed analysis} tasks of the CLPD process using two distinct experiments, which we present in \Cref{sec.eval_cr} and \Cref{sec.eval_da}. 

The candidate retrieval experiment focuses on covering a wide range of language pairs and corpora. We exclusively include in this evaluation detection methods for which source code or sufficient details for re-implementing the methods are available. For some state-of-the-art methods like CL-KGA, this is not the case, which is why we did not include them in this experiment.

The detailed analysis experiment focuses on comparing CL-OSA to state-of-the-art detection methods, some of which are not available as source code and too complex to be re-implemented. Therefore, we evaluate CL-OSA according to the protocol used in a prior study and compare our results to those reported in this study \cite{FrancoSalvadorGRB16}. The data and source code of our experiments are available at

\begin{center}
    \url{https://doi.org/10.5281/zenodo.5159398}
\end{center}

\subsection{Candidate Retrieval Evaluation}\label{sec.eval_cr}
In this evaluation, we compare CL-OSA's effectiveness in retrieving documents from five multilingual parallel corpora to four state-of-the-art CLPD methods.

\subsubsection{Datasets} Using random sampling, we selected 2,000 aligned documents from each of the following five corpora:

\textbf{PAN-PC-11 Plagiarism Corpus} \cite{PotthastEBS11}. The corpus contains instances of simulated monolingual and cross-language plagiarism that were used for evaluating plagiarism detection methods as part of the workshop series \textbf{P}lagiarism Analysis, \textbf{A}uthorship Identification, and \textbf{N}ear-Duplicate Detection (PAN). Most of the 26,939 documents in the corpus were created by extracting text from openly available books. The documents are partially interspersed with instances of simulated plagiarism that were created and obfuscated automatically or by crowdsourced workers. For the candidate retrieval evaluation, we exclusively sampled test cases from the 2,921 \textit{Spanish-English} aligned document pairs in the corpus.

\textbf{ASPEC-JE} \cite{NakazawaYUU16}. This subset of the Asian Scientific Paper Excerpt Corpus (ASPEC) contains abstracts of approx. two million research papers that were translated manually from \textit{Japanese} to \textit{English}.  

\textbf{ASPEC-JC} \cite{NakazawaYUU16}. This subset of the ASPEC corpus contains abstracts and paragraphs from the main text of research papers that were translated manually from \textit{Japanese} to \textit{Chinese}.

\textbf{JRC-Acquis} \cite{SteinbergerPWI06}. The corpus consists of legislative texts in 22 languages, which the European Union's Joint Research Centre (JRC) selected from the cumulative body of EU laws (the so called \textit{Acquis communautaire}\footnote{https://ec.europa.eu/jrc/en/language-technologies/jrc-acquis}). We sampled our test cases from the 10,000 document pairs in the \textit{English-French} subset of the corpus.

\textbf{Europarl} \cite{Koehn05}. The corpus contains transcripts of European Parliament proceedings in 21 European languages. As for JRC-Acquis, we exclusively sampled test cases from the 9,443 document pairs in the \textit{English-French} subset of the corpus.

We used the subsets of the PAN-PC-11, JRC-Acquis and Europarl corpora that Ferrero et al. \cite{FerreroABS16} pre-selected and provided for the evaluation of cross-language similarity detection methods.

Except for PAN-PC-11, all corpora contain exactly one relevant item for each query. In the PAN-PC-11 corpus, plagiarized text fragments can originate from several source documents. 

\subsubsection{CLPD Methods}
We compare CL-OSA to these methods:

\textbf{CL-ASA} implemented according to Potthast et al.~\cite[p. 9]{PotthastBSR11}. For European languages, we derived the translation probabilities from the dictionaries provided by Aker et al. \cite{AkerPPG14}. For the ASPEC corpora (EN-JA and JA-ZH), we used the program \textit{pialign} by Neubig et al. \cite{NeubigWSM11} to train the probabilities on the Tanaka corpus \cite{Tanaka01} and the TED English Chinese Parallel Corpus of Speech \cite{CettoloGF12}, respectively.

\textbf{CL-ESA} \cite{PotthastSA08} uses a comparable corpus of 20,000 Wikipedia articles, i.e., twice the upper boundary of the dimensionality interval that Potthast et al. reported achieving a good trade-off between retrieval quality and computing time \cite[p.~526f.]{PotthastSA08}.

\textbf{ConceptNet} \cite{SpeerCH17} refers to the pre-created set of vectors from  ConceptNet-Numberbatch available on  
GitHub\footnote{\url{https://github.com/commonsense/conceptnet-numberbatch}}.

\textbf{USE-ML} \cite{YangCAG20} refers to the pre-trained model Universal Sentence Encoder-Multilingual based on the Transformer architecture introduced by Yang et al. \cite{YangCAG20} and available in TensorFlow\footnote{\url{https://tfhub.dev/google/universal-sentence-encoder-multilingual/2}}. This model was trained with the Stanford Natural Language Inference corpus. 

The vectorized documents for ConceptNet and USE-ML are available in our Zenodo repository.

\begin{table*}[t!]
\centering
\caption{Detailed analysis results for entire corpus subsets.}
\resizebox{0.6\textwidth}{!}{
\begin{threeparttable}[ht!]
\label{tab.eval_da.overall}
\begin{tabular}{@{\extracolsep{4pt}}lcccccccc}
\toprule   
{} & \multicolumn{4}{c}{\textbf{Spanish-English}}  & \multicolumn{4}{c}{\textbf{German-English}}\\
 \cmidrule{2-5} 
 \cmidrule{6-9} 
 \textbf{Model} & Q & P & R & G & Q & P & R & G  \\ 
\midrule
CL-OSA  & 0.573 & 0.723 & 0.474 & 1.000 & \textbf{0.521} & 0.672 & 0.425 & 1.000\\ 
CL-KGA   & \textbf{0.620} & 0.696 & 0.558 & 1.000 & 0.520 & 0.601 & 0.460 & 1.004  \\ 
CL-VSM   & 0.564 & 0.630 & 0.517 & 1.010 & 0.414 & 0.524 & 0.362 & 1.048  \\ 
CL-ASA   & 0.517 & 0.690 & 0.448 & 1.071 & 0.406 & 0.604 & 0.344 & 1.113  \\ 
CL-ESA   & 0.471 & 0.535 & 0.448 & 1.048 & 0.269 & 0.402 & 0.230 & 1.125  \\ 
CL-C3G   & 0.373 & 0.563 & 0.324 & 1.148 & 0.115 & 0.316 & 0.080 & 1.166  \\ 
XCNN   & 0.386 & 0.738 & 0.310 & 1.189 & 0.270 & 0.664 & 0.196 & 1.174  \\ 
S2Net   & 0.514 & 0.734 & 0.440 & 1.098 & 0.379 & 0.669 & 0.304 & 1.148  \\ 
BAE   & 0.440 & 0.736 & 0.360 & 1.142 & 0.212 & 0.482 & 0.150 & 1.120  \\ 
\bottomrule
\end{tabular}

\begin{tablenotes}[online]
\item[\ding{227}] Results for methods other than CL-OSA are taken from \cite{FrancoSalvadorGRB16}.
\item[\ding{227}] \textbf{Boldface} indicates the best PlagDet score for each corpus subset.
\item [\ding{227}] Column Labels: PlagDet score (Q), Precision (P), Recall (R), Granularity (G)
\end{tablenotes}
\end{threeparttable}
}
\end{table*}

\subsubsection{Performance Measures}
Recall and the size of the candidate set are essential performance indicators for the candidate retrieval stage. A method's recall, i.e., which percentage of all source documents the method retrieves among the candidates, is critical because failing to retrieve a source prohibits detecting content that originates from that source in the subsequent detailed analysis stage. The number of candidates necessary to achieve sufficient recall strongly influences the computational effort required for the analysis.

Therefore, we assess the methods' effectiveness for retrieving candidate documents and ranking them highly (thus enabling small candidate sets) in terms of the Recall at Rank (R@k) and Average Recall at Rank (ARR) measures. For easier comparability of the methods, we report the Mean Reciprocal Rank (MRR) as a single measure, quantifying a method's overall ranking performance.   

\subsubsection{Results Candidate Retrieval}

\Cref{tab.eval_cr.results} shows the results for the candidate retrieval task. CL-OSA outperforms the other methods for all corpora, which indicates our method is least affected by the diverse topical domains of the corpora and the lexical and syntactic differences of the languages. CL-OSA is also effective in retrieving documents written in distant languages, such as Japanese and English (cf. ASPEC-JE). All other methods except Cl-ASA are significantly less effective for Japanese and English than for closer language pairs like Japanese and Chinese (cf. ASPEC-JC).

All methods exhibit a significant drop in their effectiveness for the Europarl corpus. The likely reason is that the transcripts of political proceedings in this corpus often contain boilerplate text, i.e., frequent words that do not convey additional meaning, such as \textit{parliament}, \textit{resumption of the session}, or \textit{declare}.

CL-ESA performs the poorest for all corpora. A likely reason is that the dimensionality of the concept space (20,000) is too low, although it exceeds the recommendation of Potthast et al.\cite[p.~526f.]{PotthastSA08} that 5,000 to 10,000 Wikipedia articles represents a reasonable trade-off between time and retrieval quality. In an experiment by Ashgaria et al. \cite{AsghariFMF19}, CL-ESA achieved similar results.

CL-OSA's advantage is particularly strong for the PAN-PC-11 corpus, which is designed to test plagiarism detection methods.  CL\nobreakdash-OSA outperforms the second-best method (ConceptNet) by 12.71\% in terms of MRR and 17.95\% in terms of R@1. This result shows the suitability of CL-OSA for the CLPD task.


\subsection{Detailed Analysis Evaluation}\label{sec.eval_da}
This evaluation quantifies the effectiveness of CL-OSA in aligning plagiarized text fragments and their sources at the level of characters. We compare CL-OSA's results to those of eight state-of-the-art CLPD methods reported by Franco-Salvador et al. in the most comprehensive evaluation of CLPD methods to date \cite{FrancoSalvadorGRB16}.

\subsubsection{Datasets}
In accordance with the experiments by Franco-Salvador et al. \cite{FrancoSalvadorGRB16}, we used the \textit{English-Spanish} and \textit{English-German} subsets of the PAN-PC-11 corpus \cite{PotthastEBS11}. To our knowledge, these are the only datasets that offer the necessary ground-truth information on ``plagiarized`` segments at the level of characters. Opposed to our evaluation of the candidate retrieval task (\Cref{sec.eval_cr}), for which we reused a sample of the PAN-PC-11 corpus provided by Ferrero et al. \cite{FerreroABS16}, we extracted the two cross-language subsets directly from the original corpus \cite{PotthastSEB11}.

The subsets consist of simulated cross-language plagiarism instances of different lengths embedded into topically related text. Most of the ``plagiarized`` text segments that were taken from documents in the other language were machine-translated. Additionally, hired workers obfuscated approx. 1\% of those machine-translated segments manually to increase their obfuscation and make them more challenging to detect \cite{PotthastEBS11}. 
\Cref{tab.eval_da.pan-pc_corpus_stats} summarizes the two datasets. 

\begin{table}[ht]
\caption{Overview of the German-English (DE-EN) and Spanish- English (ES-EN) subsets of the PAN-PC-11 corpus.}
\label{tab.eval_da.pan-pc_corpus_stats}

\begin{tabular}{ll}
\hline
\\
\textbf{German-English documents}               &            \\
Suspicious                                      & 251        \\ 
Source                                          & 348        \\
\textbf{Spanish-English documents}              &            \\
Suspicious                                      & 304        \\
Source                                          & 202        \\

\textbf{Plagiarism cases (DE-EN, ES-EN combined)} &         \\
Case length                                   &             \\
$\bullet$  Long cases                                  & 1,506       \\
$\bullet$   Medium cases                                & 2,118       \\
$\bullet$  Short cases                                 & 1,951       \\

Obfuscation                                   &             \\
$\bullet$   Machine translation                         & 5,142       \\
$\bullet$   Machine translation + manual obfuscation    & 433         \\
\\
\hline
\end{tabular}
\end{table}

\subsubsection{CLPD Methods}
We compare CL-OSA to eight CLPD methods evaluated by Franco-Salvador et al. \cite{FrancoSalvadorGRB16}, which cover all prominent approaches to CLPD discussed in \Cref{sec.related_work}:
\begin{itemize}
    \item Machine Translation: CL-ASA, CL-CNG (specifically cross-language character 3-grams CL-C3G), CL-VSM
    \item Corpus-based Semantics: CL-ESA
    \item Knowledge-based Semantics: CL-KGA
    \item Neural Networks: BAE, S2Net, XCNN
\end{itemize}

\subsubsection{Methodology} \label{sec.eval_da.methodology}
\begin{table*}[htb!]
\centering
\caption{Detailed analysis results by plagiarism case length.}
\resizebox{0.70\textwidth}{!}{
\begin{threeparttable}[ht!]
\label{tab.eval_da.case_length}
\begin{tabular}{@{\extracolsep{4pt}}llcccccccc}
\toprule   
{} & {} & \multicolumn{4}{c}{\textbf{Spanish-English}}  & \multicolumn{4}{c}{\textbf{German-English}}\\
 \cmidrule{3-6} 
 \cmidrule{7-10}
 \textbf{Case Length} & \textbf{Model} & Q & P & R & G & Q & P & R & G  \\ 
\midrule
& CL-OSA  & 0.366 & 0.773 & 0.240 & 1.000 & 0.331 & 0.737 & 0.214 & 1.000\\ 
& CL-KGA   & \textbf{0.406} & 0.414 & 0.398 & 1.000 & \textbf{0.366} & 0.392 & 0.347 & 1.006  \\ 
& CL-VSM   & 0.399 & 0.416 & 0.391 & 1.016 & 0.320 & 0.386 & 0.300 & 1.077  \\ 
Long & CL-ASA   & 0.411 & 0.535 & 0.375 & 1.106 & 0.339 & 0.513 & 0.299 & 1.168  \\ 
cases & CL-ESA   & 0.351 & 0.388 & 0.352 & 1.076 & 0.220 & 0.329 & 0.198 & 1.176  \\ 
(x > 5,000 chars.) & CL-C3G   & 0.299 & 0.467 & 0.269 & 1.207 & 0.090 & 0.275 & 0.064 & 1.227  \\ 
& XCNN   & 0.327 & 0.655 & 0.271 & 1.253 & 0.230 & 0.619 & 0.170 & 1.234  \\ 
& S2Net   & 0.411 & 0.587 & 0.368 & 1.145 & 0.322 & 0.589 & 0.269 & 1.212  \\ 
& BAE   & 0.369 & 0.631 & 0.314 & 1.200 & 0.178 & 0.449 & 0.127 & 1.159  \\ \midrule
& CL-OSA  & \textbf{0.317}  & 0.713 & 0.204 & 1.000 &  \textbf{0.284} & 0.659 & 0.180 & 1.000\\ 
& CL-KGA   & 0.224 & 0.224 & 0.225 & 1.000 & 0.211 & 0.231 & 0.193 & 1.000  \\ 
& CL-VSM   & 0.205 & 0.215 & 0.196 & 1.000 & 0.155 & 0.183 & 0.134 & 1.000  \\ 
Medium & CL-ASA   & 0.174 & 0.224 & 0.142 & 1.000 & 0.149 & 0.204 & 0.117 & 1.000  \\ 
cases & CL-ESA   & 0.164 & 0.174 & 0.156 & 1.000 & 0.092 & 0.113 & 0.078 & 1.000  \\ 
(700 $\leq$ x $\leq$ 5,000 chars.)& CL-C3G   & 0.131 & 0.175 & 0.105 & 1.000 & 0.041 & 0.070 & 0.029 & 1.000  \\ 
& XCNN   & 0.127 & 0.221 & 0.089 & 1.000 & 0.096 & 0.204 & 0.063 & 1.000  \\ 
& S2Net   & 0.176 & 0.240 & 0.139 & 1.000 & 0.135 & 0.217 & 0.098 & 1.000  \\ 
& BAE   & 0.148 & 0.241 & 0.107 & 1.000 & 0.072 & 0.126 & 0.051 & 1.000  \\ \midrule
& CL-OSA  & \textbf{0.054} & 0.062 & 0.048 & 1.000 & \textbf{0.053} & 0.069 & 0.043 & 1.000\\ 
& CL-KGA   & 0.012 & 0.009 & 0.021 & 1.000 & 0.011 & 0.008 & 0.018 & 1.000  \\ 
& CL-VSM   & 0.009 & 0.006 & 0.014 & 1.000 & 0.007 & 0.005 & 0.011 & 1.000  \\ 
Short & CL-ASA   & 0.006 & 0.005 & 0.009 & 1.000 & 0.006 & 0.005 & 0.009 & 1.000  \\ 
cases & CL-ESA   & 0.009 & 0.006 & 0.015 & 1.000 & 0.005 & 0.003  & 0.008 & 1.000  \\ 
(x < 700 chars.) & CL-C3G   & 0.005 & 0.004 & 0.006 & 1.000 & 0.004 & 0.003 & 0.005 & 1.000  \\ 
& XCNN   & 0.006 & 0.006 & 0.006 & 1.000 & 0.009 & 0.009 & 0.009 & 1.000  \\ 
& S2Net   & 0.008 & 0.007 & 0.010 & 1.000 & 0.008 & 0.006 & 0.010 & 1.000  \\ 
& BAE   & 0.003 & 0.003 & 0.004 & 1.000 & 0.005 & 0.004 & 0.007 & 1.000  \\ 
\bottomrule
\end{tabular}

\begin{tablenotes}[online]
\item[\ding{227}] Results for methods other than CL-OSA are taken from \cite{FrancoSalvadorGRB16}.
\item[\ding{227}] \textbf{Boldface} indicates the best PlagDet score for each corpus subset.

\item [\ding{227}] Column Labels: PlagDet score (Q), Precision (P), Recall (R), Granularity (G)
\end{tablenotes}
\end{threeparttable}
}
\end{table*}
To enable a comparison of our results to those reported in the study by Franco-Salvador et al., we adhere to the methodology of \textit{Experiment B} of the previous study \cite[p. 94ff.]{FrancoSalvadorGRB16}. 

Aligning plagiarized text segments in a document with their source segments requires the computation of similarity scores at the sub-document level. Therefore, all documents in the PAN-PC-11 subsets involved in cross-language plagiarism (both suspicious and source documents) were split into fragments. Subsequently, the evaluated CLPD methods were applied to compute the similarity scores for all possible fragment pairs. 

CL-OSA splits documents into fragments using a sliding window with a length of six sentences and a step-width of three sentences. Thus, consecutive fragments have an overlap of three sentences, which aids in identifying plagiarism that spans multiple fragments. 

We use each fragment of a suspicious document containing cross-language plagiarism as a query. For each query, we retrieve from the set of fragments obtained from the respective PAN-PC-11 subset the five fragments with the highest CL-OSA similarity score. 

To identify plagiarism that spans multiple fragments, the affected fragments need to be merged. For merging and classifying fragments as plagiarized, we used \textit{Algorithm 1} proposed by Franco-Salvador et al. \cite[p. 89]{FrancoSalvadorGRB16}. The algorithm checks if the character distance between two query fragments and their potential source fragments retrieved by a CLPD method (in our case, CL-OSA) are below a certain threshold. If so, the fragments are merged and their similarity scores accumulated. If the accumulated similarity scores of the merged fragments are above a certain threshold, the affected text segment is marked as plagiarized. We determined the best-performing thresholds for merging and classifying fragments as plagiarized via parameter tuning runs.


\subsubsection{Performance Measures}
For the detailed analysis evaluation, we use the performance measures Potthast et al. defined for this task as part of the PAN-PC competition series, i.e., Precision (P), Recall (R), Granularity (G), and PlagDet score (Q) \cite{PotthastEBS11}. Precision is the fraction of characters pertaining to a plagiarism case and the characters a method reports as plagiarized. Recall quantifies the share of all plagiarized characters a method identifies correctly. Granularity indicates whether a method reports multiple detections for a coherent plagiarism case, or yields overlapping detections, both of which are undesirable. The granularity score is in the interval $[0,1]$, with $G=1$ reflecting the best-possible case, i.e, the method reports each plagiarism case as one detection. The PlagDet score combines P, R and G into a single score
\[Q=\frac{F_1}{\log_2{(1+G)}},\] 

where $F_1$ represents the harmonic mean of Precision and Recall.

\subsubsection{Results Detailed Analysis}

\Cref{tab.eval_da.overall} shows the results of the detailed analysis evaluation on the full corpus subsets. For the Spanish-English subset, CL-OSA outperforms seven of the eight comparison methods. Only CL-KGA, which is conceptually similar to our method, achieves a slightly higher PlagDet score. For the German-English subset, CL-OSA performs marginally better than CL-KGA and significantly better than the other methods.

\Cref{tab.eval_da.case_length} presents a more fine-grained analysis of the results reported in \Cref{tab.eval_da.overall} by distinguishing the length of plagiarism cases. All methods perform better for longer cases than for shorter ones. This result is intuitive, since longer cases offer more data usable for the similarity analysis. Notably, CL-OSA performs better than all other methods in detecting short and medium cases, which are more challenging to identify. The merging algorithm described in \Cref{sec.eval_da.methodology} greatly improves CL-OSA's effectiveness for medium and long cases. The larger a plagiarism case, the more fragments will the algorithm merge, and the more likely the accumulated similarity score will be above the reporting threshold. 

\begin{table*}[ht]
\centering
\caption{Detailed analysis results by obfuscation type.}
\resizebox{0.80\textwidth}{!}{
\begin{threeparttable}[ht!]
\label{tab.eval_da.obfuscation}
\begin{tabular}{@{\extracolsep{4pt}}llcccccccc}
\toprule   
{} & {} & \multicolumn{4}{c}{\textbf{Spanish-English}}  & \multicolumn{4}{c}{\textbf{German-English}}\\
 \cmidrule{3-6} 
 \cmidrule{7-10} 
 \textbf{Obfuscation Type} & \textbf{Model} & Q & P & R & G & Q & P & R & G  \\ 
\midrule
& CL-OSA  & \textbf{0.413} & 0.506 & 0.349 & 1.000 & \textbf{0.370} & 0.475 & 0.303 & 1.000\\ 
& CL-KGA   & 0.139 & 0.158 & 0.124 & 1.000 & 0.169 & 0.207 & 0.143 & 1.000  \\ 
& CL-VSM   & 0.102 & 0.121 & 0.088 & 1.000 & 0.109 & 0.147 & 0.086 & 1.000  \\ 
Translated & CL-ASA   & 0.100 & 0.146 & 0.076 & 1.000 & 0.085 & 0.137 & 0.062 & 1.000  \\ 
manual & CL-ESA   & 0.092 & 0.107 & 0.081 & 1.000 & 0.078 & 0.122 & 0.057 & 1.000  \\ 
obfuscation & CL-C3G   & 0.072 & 0.104 & 0.054 & 1.000 & 0.042 & 0.053 & 0.035 & 1.000  \\ 
& XCNN   & 0.077 & 0.116 & 0.058 & 1.000 & 0.085 & 0.160 & 0.058 & 1.000  \\ 
& S2Net   & 0.091 & 0.141 & 0.067 & 1.000 & 0.115 & 0.173 & 0.086 & 1.000  \\ 
& BAE   & 0.085 & 0.191 & 0.055 & 1.000 & 0.088 & 0.113 & 0.072 & 1.000  \\ \midrule
& CL-OSA  & 0.584 & 0.733 & 0.485 & 1.000 & 0.533 & 0.684 & 0.434 & 1.000\\ 
& CL-KGA   & \textbf{0.660} & 0.742 & 0.595 & 1.000 & \textbf{0.556} & 0.642 & 0.493 & 1.004  \\ 
& CL-VSM   & 0.603 & 0.673 & 0.553 & 1.011 & 0.445 & 0.562 & 0.391 & 1.053  \\ 
Translated & CL-ASA   & 0.552 & 0.736 & 0.479 & 1.077 & 0.439 & 0.652 & 0.373 & 1.125  \\ 
automatic & CL-ESA   & 0.503 & 0.571 & 0.479 & 1.052 & 0.288 & 0.431 & 0.247 & 1.137  \\ 
obfuscation & CL-C3G   & 0.398 & 0.602 & 0.347 & 1.160 & 0.122 & 0.343 & 0.085 & 1.183  \\ 
& XCNN   & 0.412 & 0.791 & 0.331 & 1.205 & 0.289 & 0.715 & 0.210 & 1.191  \\ 
& S2Net   & 0.550 & 0.784 & 0.471 & 1.106 & 0.406 & 0.719 & 0.326 & 1.164  \\ 
& BAE   & 0.470 & 0.781 & 0.386 & 1.154 & 0.224 & 0.520 & 0.158 & 1.132  \\ 
\bottomrule
\end{tabular}

\begin{tablenotes}[online]
\item[\ding{227}] Results for methods other than CL-OSA are taken from \cite{FrancoSalvadorGRB16}.
\item[\ding{227}] \textbf{Boldface} indicates the best PlagDet score for each corpus subset.
\item [\ding{227}] Column Labels: PlagDet score (Q), Precision (P), Recall (R), Granularity (G)
\end{tablenotes}
\end{threeparttable}
}
\end{table*} 
  
\Cref{tab.eval_da.obfuscation} presents another breakdown of the results in \Cref{tab.eval_da.overall} according to the obfuscation applied to plagiarism cases. The results confirm that identifying manually obfuscated cases, i.e., sense-for-sense translations, is more challenging for all methods, as intended by the creators of the PAN-PC-11 dataset \cite{PotthastEBS11}. That most of the manually obfuscated cases are short further increases the difficulty of detecting them \cite{FrancoSalvadorGRB16}. CL-OSA outperforms all other methods for manually obfuscated plagiarism cases in both corpus subsets. Notably, CL-OSA's PlagDet score exceeds that of the conceptually similar method CL-KGA by a factor of 2.97 for Spanish-English and 2.19 for German-English cases. The deep semantic analysis capabilities of CL-OSA seem to provide a significant benefit for identifying these challenging plagiarism cases.

\section{Conclusion \& Future Work}
\label{sec.conclusion}

We introduced CL-OSA\textemdash a novel method that uses open knowledge graphs for cross-language plagiarism detection.
CL-OSA sets itself apart from many state-of-the-art methods by performing a deep semantic analysis of documents using entities and relationships obtained from Wikidata. Our method creates a language-independent semantic representation of documents that allows assessing the documents' similarity for many languages. CL-OSA does not require machine translation, which is a drawback of several existing methods, whose effectiveness strongly depends on the availability and quality of parallel corpora.

We evaluated CL-OSA for the candidate retrieval and detailed analysis tasks in cross-language plagiarism detection. In the candidate retrieval experiment, CL-OSA outperformed state-of-the-art CLPD methods for all five multilingual test corpora. The difference in CLPD effectiveness was most evident for the PAN-PC-11 corpus, which is tailored to the evaluation of plagiarism detection methods and includes manually translated test cases. CL-OSA's performance was unaffected by topical domains or the lack of lexical and syntactic similarities among languages. Our method also achieved excellent results for assessing the similarity of documents written in distant language pairs, such as English and Japanese, which represent a major challenge for other CLPD methods. 

In the detailed-analysis experiment, CL-OSA and the conceptually similar method CL-KGA outperform all other methods. Considering the entire test corpora, CL-KGA is slightly more effective than CL-OSA. However, our method performs significantly better than CL-KGA in detecting manually obfuscated cases of plagiarism, which are particularly challenging to identify. 

Given these results, we consider CL-OSA a promising approach to detect the highly obfuscated cross-language plagiarism we expect of researchers with strong incentives to mask wrongdoing.

In our future work, we plan to further increase the effectiveness of CL-OSA by investigating in more detail which characteristics of CL-KGA cause its performance advantage for long and automatically obfuscated cases. Moreover, we intend to optimize CL-OSA's weighting scheme for entity types. We hypothesize that using contextual information at the level of documents and fragments instead of the current boolean weighting of the term frequency will improve the selection of relevant concepts and the identification of suspicious cross-language similarity. 



\printbibliography[notkeyword=preprintref]
\end{document}